\documentclass{article}
\usepackage{spconf,amsmath,graphicx}
\usepackage{amsmath,amsfonts}
\usepackage{algorithmic}
\usepackage{algorithm}
\usepackage{array}
\usepackage{subfigure}
\usepackage{graphicx}
\usepackage{multicol}
\usepackage{multirow}
\usepackage{textcomp}
\usepackage{stfloats}
\usepackage{url}
\usepackage{verbatim}
\usepackage{graphicx}
\usepackage{cite}
\usepackage{booktabs}
\usepackage{tabularx}
\usepackage{amsfonts,amssymb}
\usepackage{multirow}
\usepackage{amsmath,amsfonts}
\usepackage{algorithmic}
\usepackage{algorithm}
\usepackage{array}
\usepackage{bbding}
\usepackage{wrapfig}
\usepackage[table]{xcolor}
\newcommand{\gr}{\rowcolor[gray]{.95}}

\title{DEHRFormer: Real-time Transformer for Depth Estimation and Haze Removal from Varicolored Haze Scenes}
\name{Sixiang Chen$^{1\dag}$, Tian Ye$^{1\dag}$, Jun Shi$^{2\dag}$, Yun Liu$^{3}$, JingXia Jiang$^{1}$, Erkang Chen$^{1,4}$, Peng Chen$^{1,4*}$.
\thanks{$^{1}$Sixiang Chen, Tian Ye, JingXia Jiang, Erkang Chen and Peng Chen are with the School of Ocean Information Engineering, Jimei University, Xiamen, China. E-mail: \{\tt\small 201921114013, 201921114031, 202021114006, ekchen, chenpeng\}@jmu.edu.cn}%
\thanks{$^{2}$Jun Shi is with the School of Information Science and Engineering, Xinjiang University, 
        Uruqmi, China.
        {\tt\small junshi2022@gmail.com}}%
\thanks{$^{3}$Yun Liu is with the College of Artificial Intelligence, Southwest University, Chongqing, China. 
        {\tt\small yunliu@swu.edu.cn}}%
\thanks{$^{4}$Erkang Chen and Peng Chen are with the Fujian Provincial Key Laboratory of Oceanic Information Perception and Intelligent Processing.}%
\thanks{This research was supported by Natural Science Foundation of Fujian Province, China (2021J01867), Natural Science Foundation of Chongqing, China (cstc2020jcyj-msxmX0324), Xiamen Municipal Bureau of Ocean Development (22CZB013HJ04).}
\thanks{$^{*}$: Corresponding author. $\dag$: Equal Contribution}
}
\address{}
\vspace{-1.5cm}
\begin{document}

\small
%
\maketitle
\begin{abstract}
Varicolored haze caused by chromatic casts poses haze removal and depth estimation challenges. Recent learning-based depth estimation methods are mainly targeted at dehazing first and estimating depth subsequently from haze-free scenes. This way, the inner connections between colored haze and scene depth are lost. In this paper,
we propose a real-time transformer for simultaneous single image Depth Estimation and Haze
Removal (DEHRFormer). DEHRFormer consists of a single encoder and two task-specific decoders. The transformer decoders with learnable queries are designed to decode coupling features from the task-agnostic encoder and project them into clean image and depth map, respectively. In addition, we introduce a novel learning paradigm that utilizes contrastive learning and domain consistency learning to tackle weak-generalization problem for real-world dehazing, while predicting the same depth map from the same scene with varicolored haze. Experiments demonstrate that DEHRFormer achieves significant performance improvement across diverse varicolored haze scenes over previous depth estimation networks and dehazing approaches.
\end{abstract}
%
%
\vspace{-0.3cm}
\section{Introduction}
\label{sec:intro}
With the development of deep learning technology, the computer vision community has entered a prosperous era~\cite{li2022ntire,yang2023aim,ye2022mutual,dong2022prior,zou2022self,chen2022msp,xie2023translating}.

Low-level vision tasks are further developed as deep learning advances~\cite{pmnet,Ye_2022_CVPR,jin2022unsupervised,chen2022snowformer,Ye_2022_ACCV,jin2022estimating,jin2023structure}.
Haze, as a common weather phenomenon, would result in severe visibility degradation, which also seriously harms high-level vision tasks, such as object detection, depth estimation, etc.
Therefore, single image dehazing, as a long-standing low-level vision task, the haze effect can be formulated by the following well-known atmosphere scattering model mathematically:
\begin{equation}
   \mathbf{I}(\mathbf{x})=\mathbf{J}(\mathbf{x}) t(\mathbf{x})+\mathbf{A(\mathbf{x})}(1-t(\mathbf{x})),
\label{eq:hazy model}
\end{equation}
where $\mathbf{I}(\mathbf{x})$ is the observed hazy image, $\mathbf{J}(\mathbf{x})$ is the clean one, $t(\mathbf{x})$ is the transmission map and $\mathbf{A(\mathbf{x})}$ stands for the global atmospheric light. Single image dehazing is a classical ill-posed problem, due to uncertain parameters: $t(\mathbf{x})$ and $A(\mathbf{x})$, while the transmission map $t(\mathbf{x})$ is a parameter associated with depth:
\begin{equation}
    t(\mathbf{x}) = e^{-\beta d(x)},
\end{equation}
where $\beta$ is the scattering coefficient of the atmosphere and $d(x)$ is the depth map.
As previously mentioned, due to the influence of the scattering particles in the atmosphere, existing widely-employed depth sensors, like LiDAR or Kinect, etc, are not reliable and robust in haze scenes.

Varicolored haze, as a more challenge ill-posed problem, offers diverse hazy conditions with vary colors. To the best of our knowledge, the varicolored haze removal is a less-touched topic in the vision community, but it is worth exploring for many applications. 


In this work, we present a new task: jointly perform depth estimation and haze removal from varicolored haze scenes.
To handle this new task, we present a novel end-to-end transformer, namely DEHRFormer, perform \textbf{D}epth \textbf{E}stimation and \textbf{H}aze \textbf{R}emoval by a unified model. Our DEHRFormer aims to tackle several long-standing but less-touch problems as follows: \textit{(i) \textbf{Varicolored haze scenes.} Compared with common haze scenes, varicolored haze is a larger collection of haze conditions with more challenging degradations. However, most existing dehazing manners often meet difficulties when trying to handle it~\cite{chen2021psd}} \textit{(ii) \textbf{Domain gap problem for varicolored dehazing.} The domain gap between real and synthetic varicolored haze domains makes networks only supervised by synthetic data hard to generalize well for the real varicolored haze images. Previous arts usually utilize complex image translation paradigm~\cite{da} or unsupervised learning based on hand-craft priors to bridge it~\cite{chen2021psd}, which are inefficient and unstable when training.} \textit{(iii) \textbf{Domain consistency problem for varicolored image depth estimation. }Estimating depth maps from haze scenes is an existing topic, but previous manners~\cite{sdde,li2018megadepth} ignore the domain consistency between clean and hazy domains, which means different depth maps may be produced from the same scene with or without haze.}

For challenging varicolored haze scenes, we introduce haze type queries in the transformer-based dehazing decoder to learn diverse varicolored haze degradations, which employ multiple head self-attention mechanism to match learnable haze queries with sample-wise degraded features, to project degraded features into the clean feature space.
In depth decoder, we further employ learnable queries as the medium to effectively capture depth information from clean features. For domain consistency problem, we present domain consistency learning to maintain the consistency of depth maps over haze and clean scenes. For domain gap problem, we introduce a novel semi-supervised contrastive learning paradigm, which explicitly exploits the knowledge from real-negative samples to boost generalization of DEHRFormer on real varicolored scenes. Moreover, we propose the first varicolored haze scene depth estimation dataset, which consists of 8000 paired data for varicolored haze removal and depth estimation tasks. We summarize the contributions as follows:
\begin{itemize}
    \item This work focuses on a novel and practical tasks: varicolored image dehazing and depth estimation. Compared with prior arts which only consider common haze removal (grayish haze scenes) or depth estimation from clean scenes, we are the first to joint consider dehazing and estimating depth maps from varicolored haze scenes in a unified way.
    
    \item We propose a real-time transformer for depth estimation and haze removal, which unifies the challenging varicolored haze removal and depth estimation to a sequence-to-sequence translation task with learnable queries, significantly easing the task pipeline.
    
    \item A semi-supervised learning paradigm is proposed to boost the generalization of DEHRFormer in the real haze domain. Furthermore, we considered the domain consistency of depth estimation over the haze and clean domains.

\end{itemize}

\vspace{-0.3cm}
\section{METHOD}


\begin{figure*}[t!]
    \centering
    \vspace{-1.2cm}
    \includegraphics[width=0.72\textwidth]{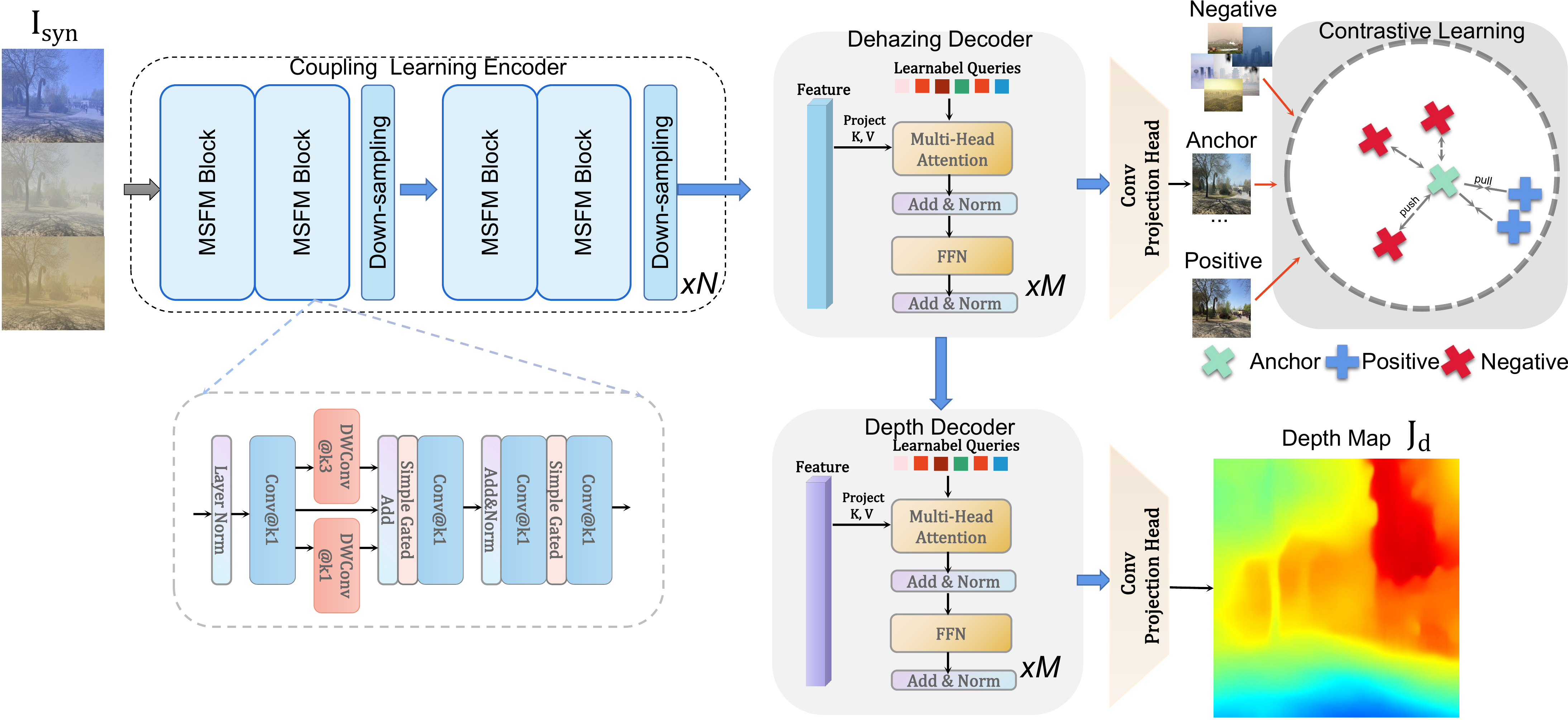}
    \caption{\small The overview of network architecture. The Coupling Learning Encoder consists of 4 scale-levels, which means that the $N$ is set as 4 in our experiments. For each encoder stage, there only two Multi-Scale Feature Modeling (MSFM) blocks to extract features for transformer decoders. For dehazing decoder and depth decoder, we set $M$ as 5 for a better trade-off between performance and model-parameters.}
    \label{fig:architecture}
\end{figure*}
\vspace{0em}

\subsection{Coupling Learning Encoder}
In our architecture, we first offer a Coupling Learning Encoder to capture features from degraded images. Different from previous approaches~\cite{valanarasu2022transweather,song2022vision}, in the encoder, we apply a CNNs-based encoder to extract coupling features of dehazing and depth estimation, which is the basis for achieving real-time efficiency for inference due to its $O(N)$ complexity compared to $O(N^{2})$ computational complexity of self-attention.
Inspired from NAFNet~\cite{chen2022simple}, the high-dimension space is crucial for extracting features. Nevertheless, it only adopts the simple 3$\times$3 depth-wise convolution to perform modeling. To enhance the coupling learning ability of features for the encoder, we propose Multi-scale Feature Modeling block (MSFM), in which multi-scale convolution is presented in high-dimension space to boost the performance of excavating coupling features of haze removal and depth estimation. As shown in Fig.\ref{fig:architecture}, given an input feature $X_{i}^{e}$, our Coupling Learning Encoder can be expressed as:
\begin{equation}
    X_{i+1}^e=\operatorname{MSFM}^{N}\left(X_i^e \in \mathbb{R}^{\frac{\mu}{2^i} \times \frac{w}{2^i} \times C_i} \downarrow\right),
\end{equation}
where the $e$ denotes the encoder, $X_{i}$ indicates the feature of $i$-th layer encoder. $H$ and $W$ mean the height and width of input image. $\downarrow$ is the down-sampling operation, we perform overlapped patch merging as follows \cite{xie2021segformer}. There are four stages in our coupling encoder.

\subsection{Task-specific Decoder For Features Decoupling}
\vspace{-0.2cm}
Motivated by DETR\cite{detr}, we attempt to use learnable queries to decode coupling features via a unified task-specific decoder. The decoding can be seen as sequence-to-sequence translation, which exploits learnable queries to translate the features from the coupling encoder. \textbf{For the dehazing decoder}, we aim to utilize the learnable haze queries $Q_{h}$ as prototypes to study varicolored degradations. Specifically, the degradation-wise type queries $Q_{h}$ are used to adaptively decode the varicolored information from encoder via a sequence-to-sequence manner. Given the coupling feature $X_{4}^{e}\in \mathbb{R}^{\frac{H}{16} \times \frac{W}{16} \times C_4}$, we feed it into our Task-specific Decoder of dehazing and reshape it into 3d sequence $X_{4}^{s}\in \mathbb{R}^{N \times C_4}$, where $N=\frac{H}{16}\times\frac{W}{16}$. We employ the linear layer to project $X_{4}^{s}$ into Key (K) and Value (V). Therefore, the decoupling can be expressed as follows self-attention:
\begin{equation}
    X_{h}^{'} =\operatorname{Softmax}\left(\frac{\mathbf{Q_{h} K}^T}{\sqrt{C_{4}}}\right) \mathbf{V},
\end{equation}
where $X_{h}^{'}$ denotes the dehazing feature decoupled from the encoder. For the decoupling feature, we then use up-sampling layer to go back to the original resolution. We add Residual Block~\cite{resnet} in each stage and have skip connections across each stage.
\textbf{For the depth decoder}, we devise the decoder to decouple the features into depth estimation space. The learnable depth queries $X_{d}$ in this decoder are leveraged to decode the depth information for various scenes. Unlike the dehazing decoder, we utilize $X_{d}$ to decouple the depth feature from the dehazing feature $X_{h}^{'}$ instead of encoder feature $X_{4}^{e}$. This can promote the network to process depth estimation from the clean feature to some extent. Therefore, our unified Task-specific decoder is serial. The overall self-attention and recover original feature size are consistent with the method of dehazing decoder.
\vspace{-0.5cm}
\subsection{Semi-supervised Contrastive Learning}
\vspace{-0.3cm}
For image restoration, conventional contrastive learning paradigm usually only exploit synthetic negative samples to boost model performance. However, real degraded samples are accessible for us, these manners ignore this point and only focus on how to boost the model performance of synthetic domain.
For bridging the gap between synthetic and real domain,
we introduce the semi-supervised contrastive learning:
\begin{equation}
\resizebox{!}{0.7cm}{
$\begin{aligned}
&\mathcal{C}\left(v, v^{+}, v^{-}\right)= \\
&-\log \left[\frac{\exp \left(\delta(v) \cdot \delta\left(v^{+}\right) \right)}{\exp \left(\delta(v) \cdot \delta\left(v^{+}\right) / \right)+\sum_{n=1}^N \exp \left(\delta(v) \cdot \delta\left(v_r^{-}\right)\right)}\right]
\end{aligned}$
}
\end{equation}
where $\delta(v),\delta(v)_{+},\delta(v)_{-}$ denote the anchor sample, positive sample, and negative sample, respectively. $\delta(\cdot)$ is the feature extraction operation by the VGG-19 network. And $N$ denotes the total number of negative samples. Our contrastive learning loss is defined as:
\begin{equation}
\mathcal{L}_{CR}=\mathcal{C}\left({J}_{syn}, J_{g t},\left\{\mathbf{ \hat{I} }_n\right\}_{n=1}^N\right)
\end{equation}
where $J_{gt}$ is the ground truth of the input image, $J_{syn}$ is the output result of DEHRFormer and $N$ is the total number of real-world varicolored haze images in a single batch.

In our semi-supervised paradigm, we exploit real-world hazy samples as negative samples, and predicated results as anchor samples. Different from previous contrastive learning manners, we leverage the set of real hazy images by all varicolored types as the negative samples. The positive sample guides our DEHRFormer to mine clean knowledge by the feature space, while real-negative samples enhance the discriminative knowledge of our network for diverse varicolored haze images. Our semi-supervised learning paradigm provides a lower bound to limit the output of DEHRFormer away from real-world negative samples, which enhances the generalization of our model on the real domain.

\vspace{-0.2cm}
\subsection{Domain Consistency Learning}
Popular depth estimation networks that only learn from the clean domain usually meet failures in haze scenes. Vice versa, the model that only learns knowledge from the haze domain would meet the generalization decline problem in the clean domain, which is not obviously neglectable. To achieve the generalization consistency between both domains, we present an additional constraint, which enforces our DEHRFormer to predict the same depth maps from the same scenes with or without hazy degradations.
Let's denote $J_{gt}^{d}$ and $J_{clean}^{d}$ as the ground-truth depth map and the depth map predicted from the clean scene by our network. The constraint to perform domain consistency learning can be introduced as follows:

\begin{equation}
    \mathcal{L}_{DC} = D(J_{gt}^{d},J_{clean}^{d})
\end{equation}
where $D(\cdot)$ denotes the Norm-based function to measure distance. 

\vspace{-0.1cm}
\subsection{Loss Functions}
We use Charbonnied loss~\cite{charbonnier1994two} as our reconstruction loss:
\begin{equation}
    {\mathcal{L}}_{\text {char }}=\frac{1}{N} \sum_{i=1}^N \sqrt{\left\|X^i-Y^i\right\|^2+\epsilon^2},
\end{equation}
where $X^{i}$ and $Y^{i}$ denote the predicted results and corresponding ground-truth. The constant $\epsilon$ is empirically set to $1e^{-3}$ for all experiments.
Our overall loss functions can be formulated as follows:
\begin{equation}
\mathcal{L} = \lambda_{1}\mathcal{L}_{char}(J^{d}_{gt},J_{syn}^{d})+\lambda_{2}\mathcal{L}_{char}(J_{gt}^{h},J_{syn}^{h}) + \lambda_{3}\mathcal{L}_{CR} + \lambda_{4}\mathcal{L}_{DC},
\end{equation}
where $J^{\left\{d,h\right\}}_{syn}$ and $J^{\left\{d,h\right\}}_{gt}$ are the estimated depth map and dehazing image, and ground-truth of depth map and haze image, respectively. $\lambda_{1},\lambda_{2}, \lambda_{3}$ and $\lambda_{4}$ are set to 1, 1, 0.5 and 1 in our all experiments.

\section{Experiments}
\noindent\textbf{Implementation Details.} 
We implement our framework using PyTorch with a RTX3090 GPU. We train our model 200 epoch with the patch size of 256$\times$256. We adopt Adam optimizer, its initial learning rate is set to $2\times 10^{-4}$, and we employ CyclicLR to adjust the learning rate. The initial momentum is set to 0.9 and 0.999. For data augmentation, we apply horizontal flipping and randomly rotate the image to 0,90,180,270 degrees.

\noindent\textbf{Datasets.} 
To facilitate the development of this task, we propose the first varicolored haze scene depth estimation
dataset, which includes 8000 paired data, named varicolored haze removal and depth estimation (VHRDE) dataset. We utilize 6,000 paired haze image from VHRDE for training and 2,000 paired data from VHRDE for testing. For real-world varicolored hazy samples, we utilize 2,000 real hazy images from the URHI (Unannotated Real Hazy Images) dataset~\cite{SOTS} for semi-supervised training and 1,000 real hazy samples for testing by No-reference image quality assessment.

\noindent\textbf{Compared with SOTA Methods}. 
We conduct extensive experiments to demonstrate the superiority of our algorithm compared to previous SOTA dehazing methods and depth estimation methods. For varicolored haze removal, we compared with DCP~\cite{dcp}, GDCP~\cite{GDCP}, PSD~\cite{chen2021psd}, SDDE\cite{sdde}, PMNet~\cite{ye2021perceiving} and NAFNet~\cite{chen2022simple}. We retrain the DL-based model on our proposed varicolored haze training set and perform inference to ensure a fair comparison. We use PSNR and SSIM to compare the performance of dehazing quantitatively. We can observe that the proposed DEHRFormer achieves the best results on PSNR and SSIM metrics in Table.\ref{hazeresults}. Compared to the second best approach NAFNet~\cite{chen2022simple}, we exceed the 0.41dB and 0.1 on PSNR and SSIM. We also present the visual comparison with previous SOTA methods in Fig.\ref{fig:syn}. It can be seen that our method can remove the varicolored haze thoroughly, while the previous methods still have various residual haze. Also, as shown in Table.~\ref{hazeresults}, we employ the well-known no-reference image quality assessment indicator to highlight our merits in real-domain, i.e., NIMA~\cite{talebi2018nima}, which predicts aesthetic qualities of images. Fig.\ref{fig:real} presents the different dehazing results in real-world images, our method obtains the best results of removing all varicolored haze compared to other algorithms.

\begin{table}[!h]
\vspace{-0.6cm}
\setlength{\belowdisplayskip}{0pt}
\setlength{\abovedisplayskip}{0pt}
 \setlength{\abovecaptionskip}{0em} 
 \setlength{\belowcaptionskip}{0em}

\centering
\caption{\small Dehazing results on the proposed VHRDE dataset and real-world dataset. Bold and underline indicate the best and second best metrics. }\label{hazeresults}
\resizebox{5cm}{!}{
\renewcommand\arraystretch{1.1}

\begin{tabular}{l|cccccc}
\toprule[1.2pt]

{Method} & \multicolumn{2}{c}{PSNR$\uparrow$}  & \multicolumn{2}{c}{SSIM$\uparrow$} & \multicolumn{2}{c}{NIMA$\uparrow$}\\
\hline (TPAMI'10)DCP~\cite{dcp} & \multicolumn{2}{c}{13.19}  &  \multicolumn{2}{c}{0.732} & \multicolumn{2}{c}{-}\\
(CVPR'16)GDCP~\cite{GDCP} & \multicolumn{2}{c}{15.34} & \multicolumn{2}{c}{0.756}  & \multicolumn{2}{c}{-}\\
(ICRA'20)SDDE~\cite{sdde} &\multicolumn{2}{c}{20.82} & \multicolumn{2}{c}{0.768}  &\multicolumn{2}{c}{3.3203} \\
(CVPR'21 Oral)PSD~\cite{chen2021psd} & \multicolumn{2}{c}{14.12} & \multicolumn{2}{c}{0.744} & \multicolumn{2}{c}{\underline{3.4856}}\\
(ECCV'22)NAFNet~\cite{chen2022simple} & \multicolumn{2}{c}{\underline{23.01}} & \multicolumn{2}{c}{\underline{0.866}} & \multicolumn{2}{c}{3.4692}\\
(ECCV'22 Oral)PMNet~\cite{ye2021perceiving} & \multicolumn{2}{c}{22.54} & \multicolumn{2}{c}{0.845} & \multicolumn{2}{c}{3.4415} \\
\hline\hline \gr DEHRFormer & \multicolumn{2}{c}{$\mathbf{23.42}$} & \multicolumn{2}{c}{$\mathbf{0.876}$}& \multicolumn{2}{c}{$\mathbf{3.7556}$} \\ 
\bottomrule[1.2pt]
\end{tabular}}
\end{table}



For the depth estimation,  We use the most common depth estimation metrics~\cite{yang2021transformers} to quantitatively measure the performance of our model in depth estimation, including root mean square error (RMSE), Abs relative error, and accuracy $\delta_{1}$, $\delta_{2}$, $\delta_{3}$~\cite{sdde}. For a fairer and more diverse comparison, for the dehazing algorithms ~\cite{dcp}~\cite{nld}~\cite{chen2021psd}~\cite{ye2021perceiving}~\cite{chen2022simple}, we first perform the dehazing manner and then use the depth estimation model~\cite{li2018megadepth} to acquire the depth map. For SDDE~\cite{sdde} and our DEHRFormer, the depth map is directly obtained from the haze map. The quantitative metrics are presented in Table.\ref{depthresults}. We found that our framework achieves the best results on five metrics. It is worth mentioning that the paradigm of dehazing first and then depth estimation does not perform well, due to the gap between the dehazing image and the clean image. This has a significant impact on clean-to-depth networks. Our method can explicitly extract the relationship between haze and depth and facilitate depth estimation directly from haze images. The visual comparison is presented in Fig.\ref{fig:syn}. It can be seen from Fig.\ref{fig:syn} that the predicted depth map better reflects the real structure scene, and the transition of details is smoother than the SOTA approaches. We show the inference time cost\footnote{Worth noting that we compare DEHRFormer with other Dehazing-DepthEstimation pipelines and single-stage manner, (i.e, SDDE) for a fair comparison. And the time reported in the table corresponds to the time taken by each model or pipeline feed forward an image of dimension $512\times512$ during the inference stage. We perform all inference testing on an RTX3090 GPU for a fair comparison. Notably, we utilize the \textit{torch.cuda.synchronize()} API function to get accurate feed forward run-time.} in Table.\ref{depthresults}. It can be seen that DEHRFormer attracts real-time performance in the inference stage and surpass the previous SDDE method or dehazing-to-depth approaches.

\begin{figure}[h!]
\vspace{0cm}
 \setlength{\abovecaptionskip}{-0.5em} 
 \setlength{\belowcaptionskip}{-0.5em}
 
    \centering
    \includegraphics[width=0.5\textwidth]{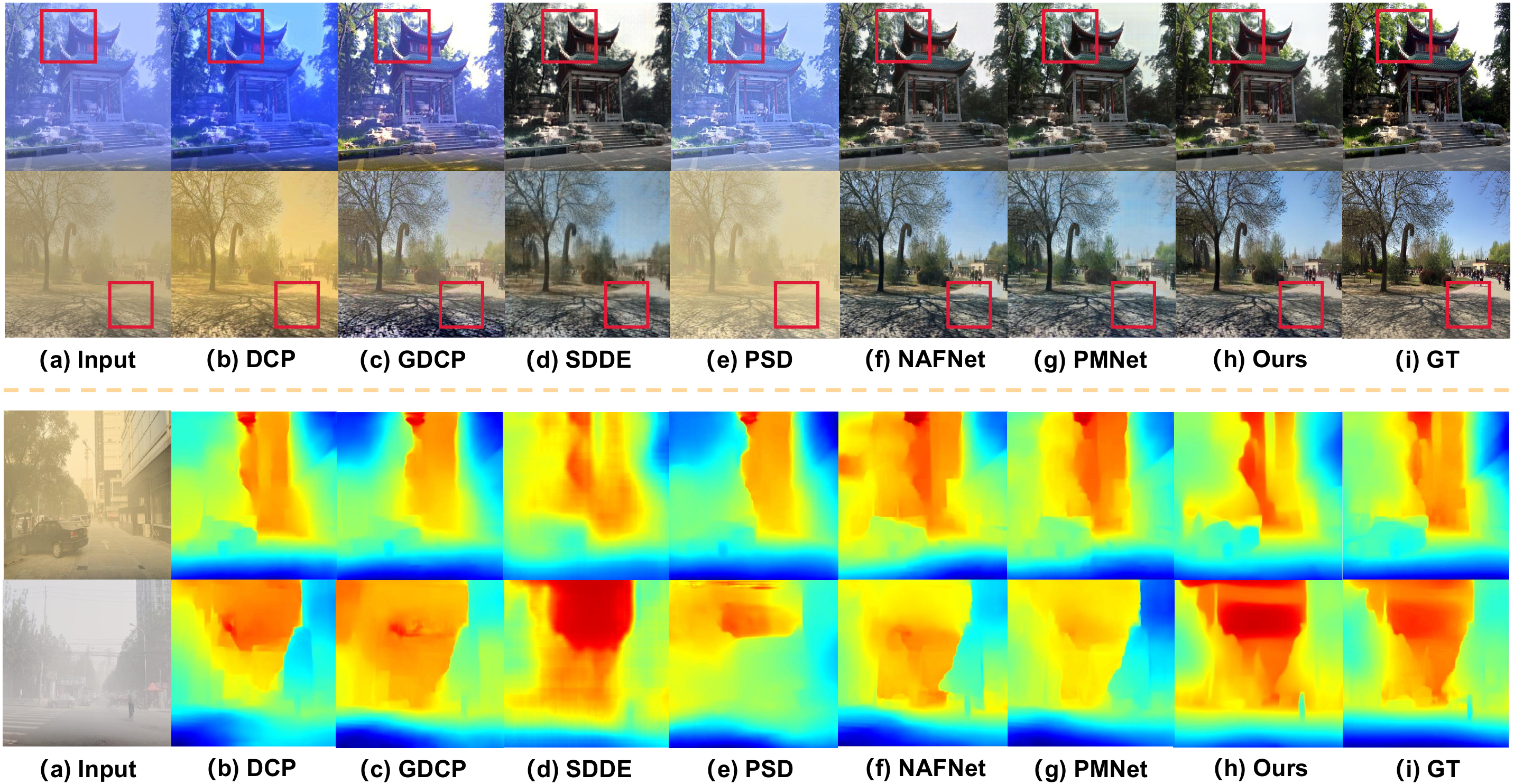}
    \caption{\footnotesize{Visual comparison of dehazing and depth estimation on the proposed VHRDE testing set.}}
    \label{fig:syn}
\end{figure}

\begin{table}[!h]
\vspace{-0.8cm}
 \setlength{\abovecaptionskip}{-0.3em} 
 \setlength{\belowcaptionskip}{-0.5cm}
 
\centering
\caption{\small{Depth Estimation results on the proposed VHRDE datasets. Bold and underline indicate the best and second best metrics.}}\label{depthresults}
\resizebox{8.5cm}{!}{
\renewcommand\arraystretch{1.1}

\begin{tabular}{l|cccccc}
\toprule[1.2pt]

{Method} & {RMSE$\downarrow$}  & {Abs Rel$\downarrow$} & $\delta_{1}\uparrow$ & $\delta_{2}\uparrow$& $\delta_{3}\uparrow$&Inf. Time(in s)\\
\hline (TPAMI'10)DCP~\cite{dcp}+MegaDepth & {0.313}  &  {0.79} & 0.289 & 0.492 & 0.633 & -\\
(CVPR'16)GDCP~\cite{GDCP}+MegaDepth& {0.305} & \underline{0.681} & 0.314 & \underline{0.528} & 0.670 & -\\
(ICRA'20)SDDE~\cite{sdde} & ${0.299}$ & ${0.688}$ &${0.303}$ & ${0.526}$ & ${0.678}$ & 0.219 \\
(CVPR'21 Oral)PSD~\cite{chen2021psd}+MegaDepth & {0.324} & {0.765} & \underline{0.324} & 0.483 & 0.634 & 0.372\\
(ECCV'22)NAFNet~\cite{chen2022simple}+MegaDepth & \underline{0.298} & {0.720} & 0.293 & 0.513 & \underline{0.677} & \underline{0.135}\\
(ECCV'22 Oral)PMNet~\cite{ye2021perceiving}+MegaDepth & {0.321} & {0.975} & 0.271&0.453 & 0.580 & 0.236 \\
\hline\hline \gr DEHRFormer & $\mathbf{0.286}$ & $\mathbf{0.640}$ &$\mathbf{0.324}$ & $\mathbf{0.552}$ & $\mathbf{0.695}$ & $\mathbf{0.034}$\\ 
\bottomrule[1.2pt]
\end{tabular}}
\end{table}




\vspace{-0.8cm}
\section{Ablation Study}
\vspace{-0.3cm}
For ablation studies, we follow the basic settings presented above and conduct experiments to demonstrate the effectiveness of the components of our proposed comprehensive manner. Next, we analyse the influence of each element individually.

\noindent\textbf{Improvements of Learnable Queries.} This part aims to demonstrate the effectiveness of proposed learnable queries in the Task-specific Decoder. We present the results in Table.\ref{query ab}. We observe that learnable queries can facilitate the decoder adaptively decouples the information we need from the coupling encoder via sequence-to-sequence. In addition, we notice that the number of learnable queries also affects the performance of the decoupling decoder. 
\begin{figure}[t!] 
\vspace{-0.8cm}
 \setlength{\abovecaptionskip}{-3cm} 
 \setlength{\belowcaptionskip}{-1em}
 
    \centering
    \includegraphics[width=0.5\textwidth]{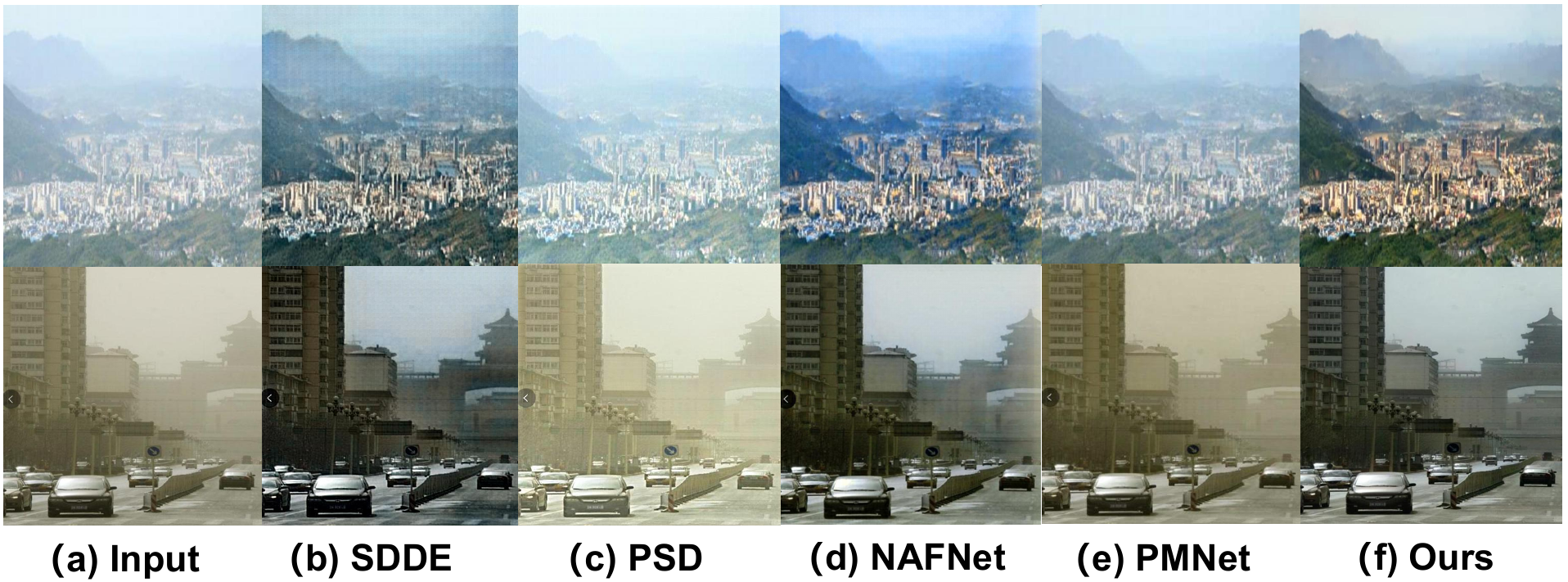}
    \caption{\small{Visual comparison of dehazing on the real-world hazy images.}}
    \label{fig:real}
\end{figure}

\begin{table}[!h]
\vspace{-0.5cm}
 \setlength{\abovecaptionskip}{-0.1cm} 
 \setlength{\belowcaptionskip}{-2.5cm}
 
\centering
\caption{\small Ablation Study on The Learnable Queries}\label{query ab}

\resizebox{4cm}{!}{
\renewcommand\arraystretch{1.1}
\begin{tabular}{lccccc}
\toprule[1.2pt]

{Method} & \multicolumn{2}{c}{RMSE$\downarrow$}  & \multicolumn{2}{c}{PSNR$\uparrow$}\\
\hline

w/o learnable queries & \multicolumn{2}{c}{{{0.301}}} & \multicolumn{2}{c}{{22.62}} \\ 
w 24 learnable queries & \multicolumn{2}{c}{{{0.295}}} & \multicolumn{2}{c}{{22.86}} \\ 
w 64 learnable queries & \multicolumn{2}{c}{{{0.291}}} & \multicolumn{2}{c}{{23.01}} \\ 
DEHRFormer & \multicolumn{2}{c}{$\mathbf{0.286}$} & \multicolumn{2}{c}{$\mathbf{23.42}$} \\ 
\bottomrule[1.2pt]
\end{tabular}}
\end{table}

\noindent\textbf{Benefits of Semi-supervised Contrastive Learning}. To boost the generalization of our model in the real-domain, we propose Semi-supervised Contrastive Learning. We tend to verify the gains of Contrastive Learning in this part.
We found that real-domain negative samples can enhance the generalization of our model. It is worth mentioning that although synthetic samples can slightly improve our metrics on synthetic datasets, the generalization ability on real-world datasets drops significantly. We also explored the effect of the ratio of positive and negative samples on model performance. We found that more negative samples can facilitate the model to exploit negative information, we only choose the 1:1 ratio to conduct the experiments due to the best trade-off between performance and graphics memory.

\begin{table}[!h]
\centering
\vspace{-0.5cm}
\caption{\small{Ablation Study on the Semi-supervised Contrastive Learining}}\label{cl ab}
 \setlength{\abovecaptionskip}{-0.8em} 
 \setlength{\belowcaptionskip}{-0.8cm}
\resizebox{3.5cm}{!}{
\renewcommand\arraystretch{1.1}
\begin{tabular}{lccccc}
\toprule[1.2pt]
{Method} & \multicolumn{2}{c}{NIMA$\uparrow$}  & \multicolumn{2}{c}{PSNR$\uparrow$}\\
\hline
w/o CL  & \multicolumn{2}{c}{{{3.4345}}} & \multicolumn{2}{c}{{22.79}} \\ 
CL w SS & \multicolumn{2}{c}{{{3.4492}}} & \multicolumn{2}{c}{${23.51}$} \\ 
CL w 1:5 & \multicolumn{2}{c}{{{3.8144}}} & \multicolumn{2}{c}{${23.55}$} \\ 
CL w 1:10 & \multicolumn{2}{c}{{$\mathbf{3.8832}$}} & \multicolumn{2}{c}{$\mathbf{23.69}$} \\ 
DEHRFormer & \multicolumn{2}{c}{${3.7556}$} & \multicolumn{2}{c}{{23.42}} \\ 
\bottomrule[1.2pt]
\end{tabular}}
\end{table}
 
\noindent\textbf{Effectiveness of Domain Consistency Learning}. To demonstrate the gain of proposed Domain Consistency Learning (DCL), we remove the domain consistency learning and observe the effect on estimating clean image depth maps directly from haze maps. From Table.\ref{DC ab}, We believe that Domain Consistency Learning can maintain the consistency between the depth maps of haze and clean scenes. 

\vspace{-0.6cm}
\begin{table}[!h]
\centering
\caption{\small{Ablation Study on The Domain Consistency Learning}}\label{DC ab}
\vspace{0cm}
\setlength{\belowdisplayskip}{0pt}
\setlength{\abovedisplayskip}{0pt}
 \setlength{\abovecaptionskip}{-0.2em} 
 \setlength{\belowcaptionskip}{-0.7cm}
 
\resizebox{4cm}{!}{
\renewcommand\arraystretch{1.1}
\begin{tabular}{lccccc}
\toprule[1.2pt]

{Method} & \multicolumn{2}{c}{RMSE$\downarrow$}  & \multicolumn{2}{c}{Abs Rel$\downarrow$}\\
\hline

w/o DCL & \multicolumn{2}{c}{{{0.299}}} & \multicolumn{2}{c}{{0.651}} \\ 
DEHRFormer & \multicolumn{2}{c}{$\mathbf{0.286}$} & \multicolumn{2}{c}{$\mathbf{0.640}$} \\ 
\bottomrule[1.2pt]
\end{tabular}}
\vspace{-0.5cm}
\end{table}
\section{CONCLUSIONS}
In this work, we propose a novel real-time transformer to tackle a new task: depth estimation and haze removal from varicolored haze scenes. Moreover, we present a semi-supervised contrastive learning paradigm for the domain gap problem to achieve domain adaptation in real-world haze scenes. To maintain depth estimation performance in clean scenes, we propose domain consistency learning to simultaneously enforce network learns from hazy and clean domains. Extensive experiments on synthetic and natural varicolored haze data demonstrate the superiority of our DEHRFormer.

\vfill\pagebreak

\bibliographystyle{IEEEbib}
\footnotesize{\bibliography{strings,refs}}

\end{document}